\documentclass[conference]{IEEEtran}
\IEEEoverridecommandlockouts
\usepackage{cite}
\usepackage{amsmath,amssymb,amsfonts}
\usepackage{algorithmic}
\usepackage{graphicx}
\usepackage{subcaption}
\usepackage{multicol}
\usepackage{lipsum}
\usepackage{pgfplotstable}

\def\BibTeX{{\rm B\kern-.05em{\sc i\kern-.025em b}\kern-.08em
    T\kern-.1667em\lower.7ex\hbox{E}\kern-.125emX}}
\begin{document}

\title{Salience Biased Loss for Object Detection \\
		 in Aerial Images \\
}

\author{\IEEEauthorblockN{Peng Sun \qquad Guang Chen \qquad Guerdan Luke\qquad Yi Shang}

\IEEEauthorblockA{
University of Missouri-Columbia\\
\tt\small \{ps793@mail, gcgrf@mail, lmg4n8@mail, shangy@\}.missouri.edu}
}

\maketitle

\begin{abstract}
    Object detection in remote sensing, especially in aerial images, remains a challenging problem due to low image resolution, complex backgrounds, and variation of scale and angles of objects in images. In current implementations, multi-scale-based and angle-based networks have been proposed and generate promising results with aerial image detection. In this paper, we propose a novel loss function, called Salience Biased Loss (SBL), for deep neural networks, which uses  salience information of the input image to achieve improved performance for object detection. Our novel loss function treats training examples differently based on input complexity in order to avoid the over-contribution of easy cases in the training process. In our experiments, RetinaNet was trained with SBL to generate an one-stage detector, SBL-RetinaNet. SBL-RetinaNet is applied to the largest existing public aerial image dataset, DOTA. Experimental results show our proposed loss function with the RetinaNet architecture  outperformed other state-of-art object detection models by at least 4.31 mAP, and RetinaNet by 2.26 mAP with the same inference speed of RetinaNet. 
\end{abstract}


\section{Introduction}
    In recent years, deep neural networks have been applied to many areas and have achieved huge success in different domains such as image classification\cite{krizhevsky2012imagenet,simonyan2014very,he2016deep}, object detection\cite{girshick2014rich,ren2015faster,he2017mask,lin2017feature,redmon2017yolo9000,dai2016r,shrivastava2016training,liu2016ssd}, and remote sensing\cite{chen2017automatic,tang2017vehicle,sommer2017deep,yang2018automatic}. Although the past decade has brought many advances in object detection, it remains a critical and challenging problem. For instance, CNNs have been applied to image classification problems in ImageNet\cite{deng2009imagenet} and surpassed the error rate of human vision ability; however, the best-performing object detection model on the COCO dataset\cite{lin2014microsoft} only achieved around 40 mAP even when the IoU of the ground truth box and predicted box is just 0.5. With the performance improvements of DNN models, the inference time of the model is another critical metric to be evaluated. Therefore, two types of object detectors are popular in the computer vision domain: one-stage and two stage detectors. The current state-of-the-art object detectors are based on a two-stage, proposal-driven mechanism. In general, the first stage generates the potential location of each target object using a R-CNN\cite{girshick2014rich}, while the second stage classifies each candidate location as foreground or background and adjusts the coordinate of each candidate using a CNN. However, in real applications, the inference and training time is usually slow since the networks are complicated. In another approach, a one stage detector has had a simpler architecture while achieving similar accuracy. One stage detectors, in general, will apply over regular and dense sampling of object scales, locations, and aspect ratios, such as YOLO\cite{redmon2017yolo9000}, SSD\cite{liu2016ssd}, and RetinaNet\cite{lin2017focal}. Each of these demonstrates promising results with faster speed, a simpler network, and similar accuracy of two-stage object detectors. RetinaNet\cite{lin2017focal} even outperforms one of the best two-stage detectors, Faster R-CNN\cite{ren2015faster}, with a relative 4.0 mAP improvement in COCO data\cite{lin2014microsoft}.
    
	\begin{figure}[h!]
  	\centering
   	\begin{subfigure}[b]{0.4\linewidth}
    	\includegraphics[width=\linewidth]{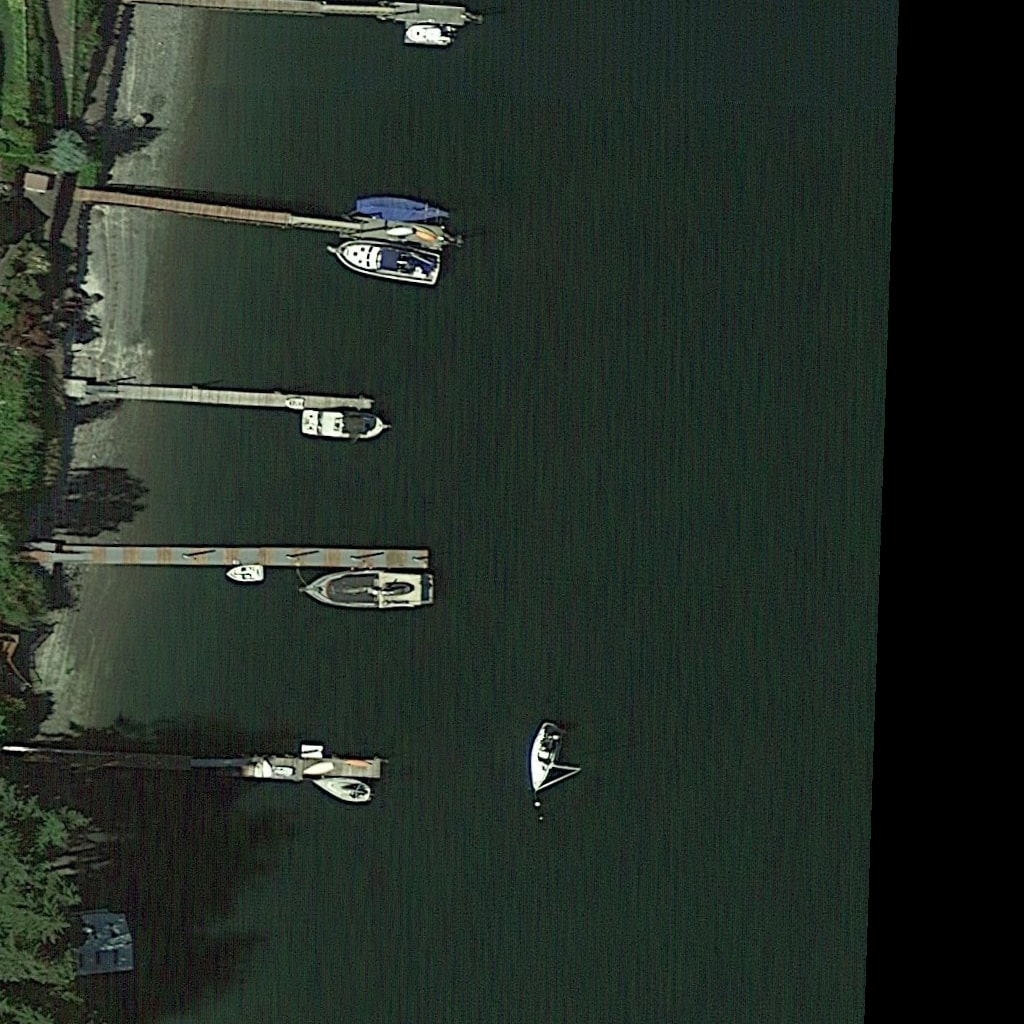}
   	\end{subfigure}
  	\begin{subfigure}[b]{0.4\linewidth}
    	\includegraphics[width=\linewidth]{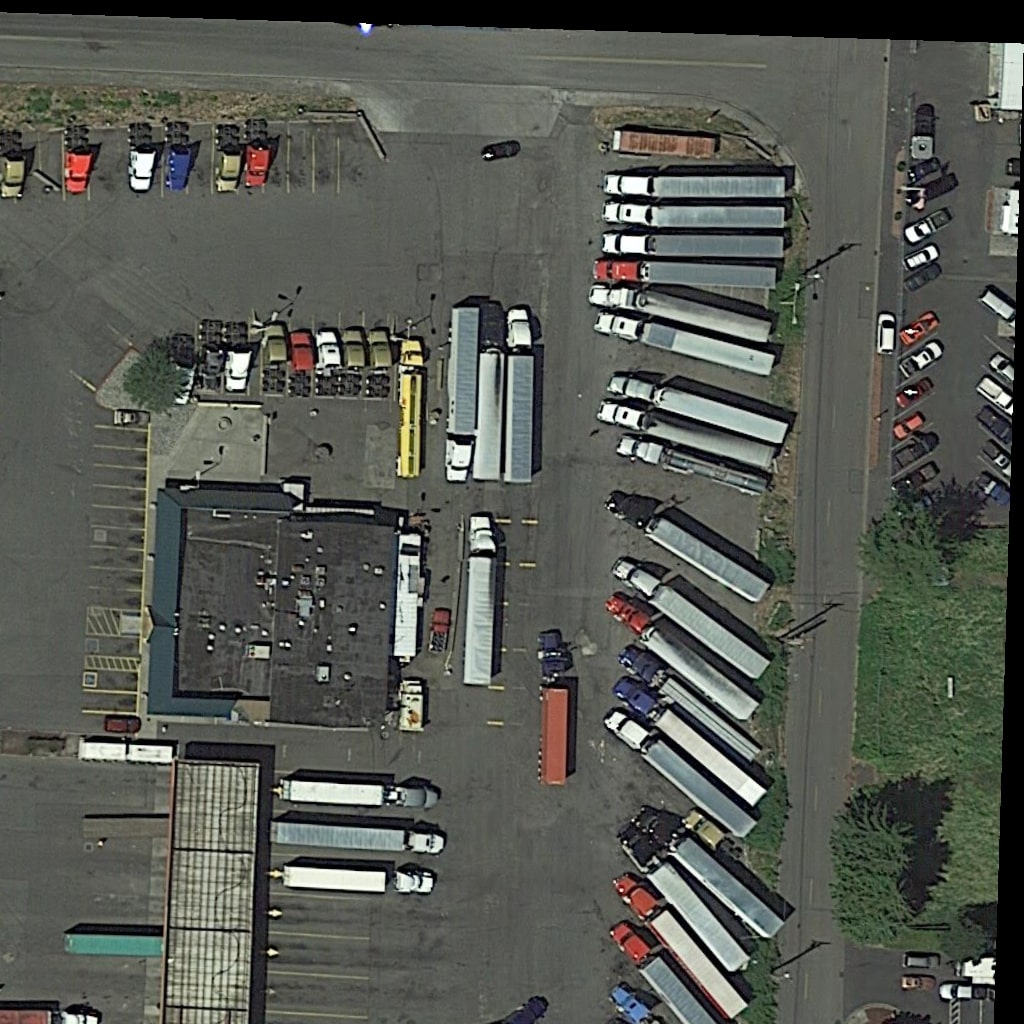}
  	\end{subfigure}
  	   	\begin{subfigure}[b]{0.4\linewidth}
    	\includegraphics[width=\linewidth]{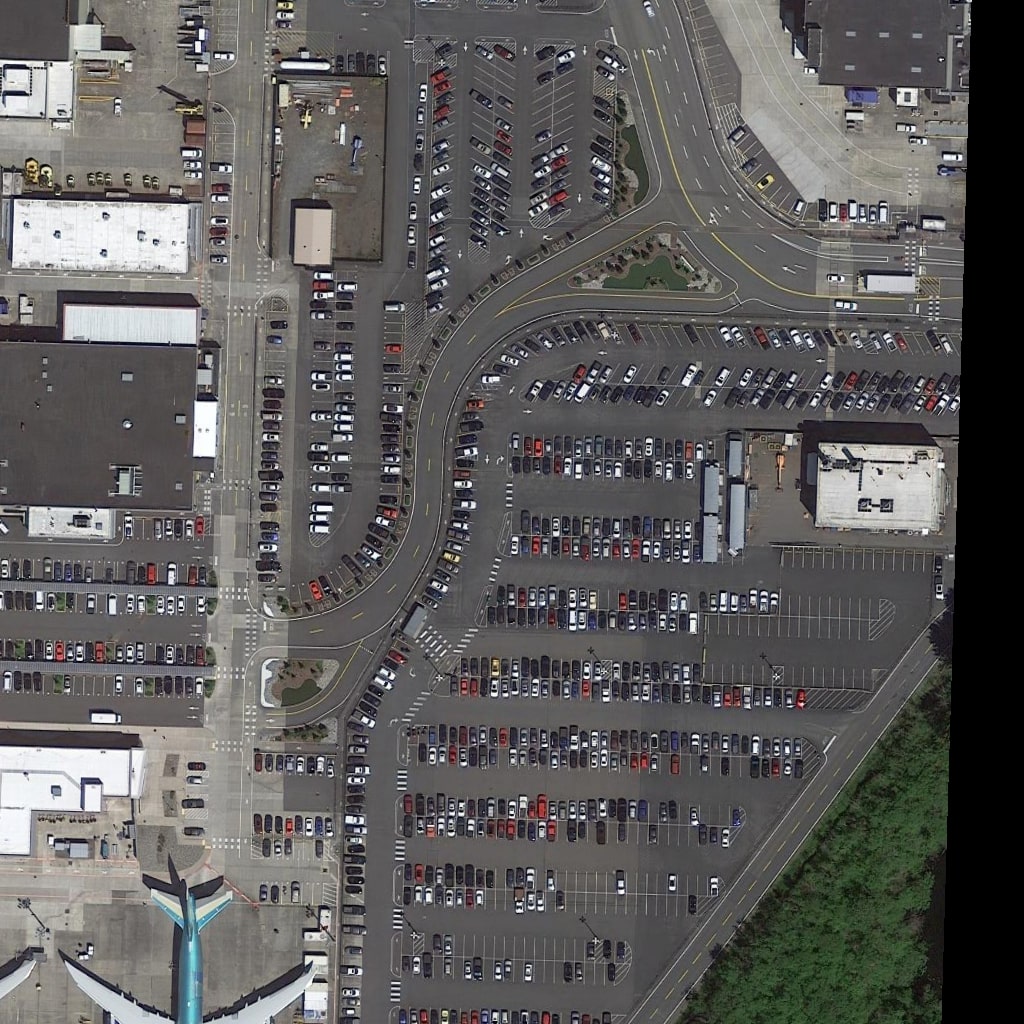}
   	\end{subfigure}
  	\begin{subfigure}[b]{0.4\linewidth}
    	\includegraphics[width=\linewidth]{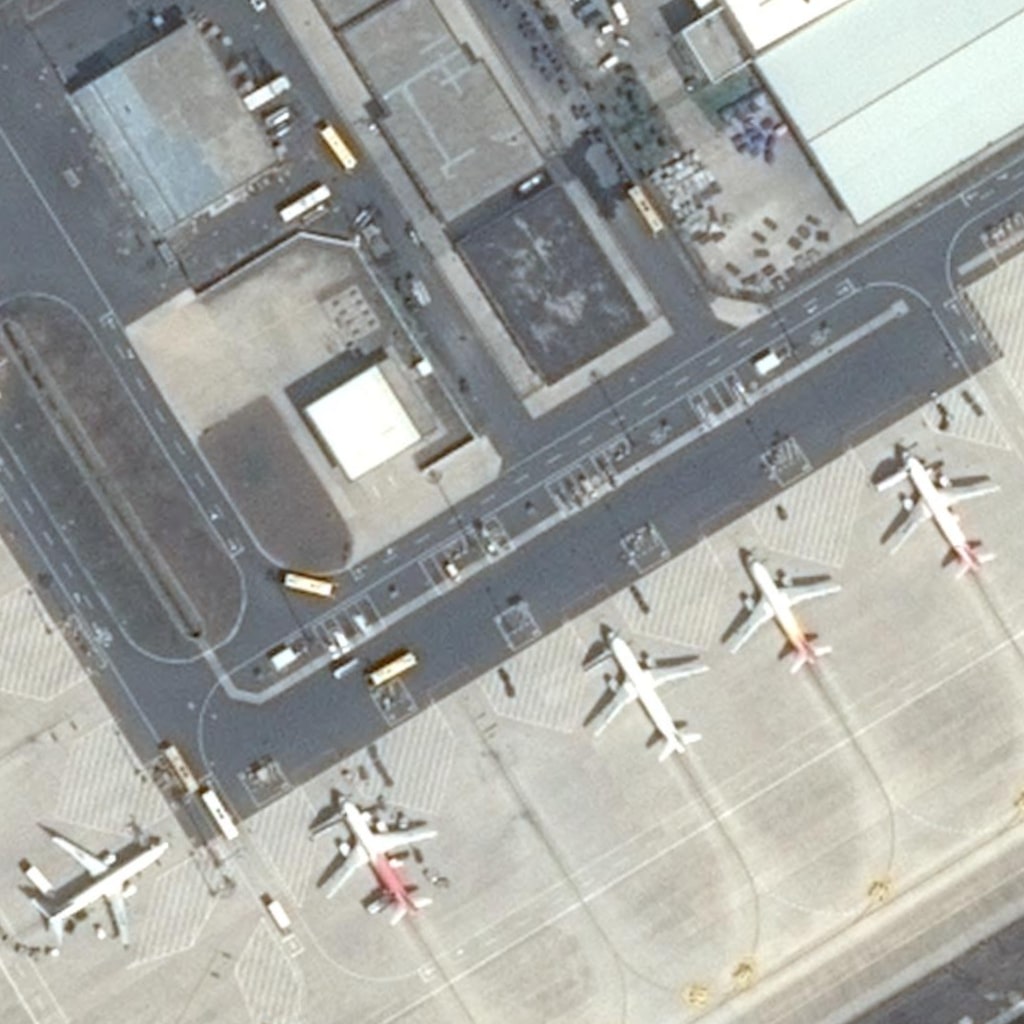}
  	\end{subfigure}
  	\caption{Sample Image of DOTA. It demonstrates the variation of scale and angles of objects in aerial images. Harbor, Plane, Small Vehicle, Large Vehicle are the objects that need to be detected.}
  	\label{fig:sample data}
	\end{figure}

    With the development and success of DNNs, deep learning has been applied to the sensor data domain in the past several years, including bio-sensors\cite{sun2016ada,bernstein2017using} and remote sensors\cite{chen2017automatic,tang2017vehicle,sommer2017deep,xia2018dota,yang2018automatic}. However, sensor data has its own character which is different from conventional object detection datasets. For example, even though the aerial images collected by remote sensors look like natural images and all the objects can be recognized by a human, there are 3 characteristics of objects in aerial images, as shown in Fig.1. (1) Objects in aerial images often appear with arbitrary orientations depending on the perspective of sensor platforms. (2) The scale variations of objects in aerial images are much larger than conventional image datasets, and many small objects are crowed in aerial images. (3) The background of some aerial images is clear and simple, while others have a more complex background. These characteristics are not typical of images in conventional object detection datasets. Based on these properties of aerial images, object detection in aerial images is a challenging problem in the computer vision and remote sensing domains. Moreover, in the real applications of aerial images, inference time is an important evaluation of models because sensor devices are dynamic capture.  
    
    Here is the problem: How to get better results for aerial images using object detection algorithms while decreasing latency speed? Based on the properties of aerial images, recent work demonstrates rotation-based\cite{li2018multiscale,azimi2018towards} and multi-scale-based networks\cite{li2018multiscale,li2018r} which focus on solving the first two challenges found in aerial images. These models demonstrate significant improvements in aerial image object detection. However, many of these contributions adopt a more complicated network architecture which may reduce inference speed. In this paper, we propose a novel loss objective function, Salience Biased Loss, that can be used to address the third challenge found in aerial images. We improve the performance of state-of-the-art object detectors while retaining the same inference time. Our new loss function treats all input images differently, focusing training on a set of complicated images that prevents the vast number of easy cases from overwhelming the converge of cross-entropy loss of the model during the training phase. In order to apply this novel loss function in an experiment, we modify a simple dense one-stage detector, RetinaNet\cite{lin2017focal}, to generate SBL-RetinaNet. In order to evaluate the performance of the model with our novel loss function, two public datasets were used, DOTA\cite{xia2018dota}, (a large-scale object detection in aerial images) and LBAI\cite{liu2018performance} (Little Birds in Aerial Imagery). Experiments show that our proposed loss function in SBL-RetinaNet outperforms other state-of-the-art object detectors. It outperforms RetinaNet with post-tuning\cite{lin2017focal} by 2.26 mAP on DOTA data and yields a 1.31\% improvement on LBAI data. 
    
    The rest of the paper is organized as follows. Section II introduces related work of object detection, with an emphasis on the techniques related to object detection on aerial images. Section III presents the theory of Salience Biased Loss and the architecture of SBL-RetinaNet. Section IV describes the ablation study and experimental results on DOTA and final results on LIBAI. Finally, Section V summarizes the paper, conclusion, and main contribution.
\section{Related Work}		
Two major approaches are popular in the object detection domain. The dominant approach in modern object detection is based on a two-stage approach which generates a set of proposed targets and detects the bounding box and label for each proposed region. The second approach is using one-stage model applied over regular and dense sampling of object scales, locations, and aspect ratios to generate the location and label for each target object. Both of these approaches can be used in aerial image object detection; however, due to the special character of aerial images, a more suitable design of the experiment is necessary.

	\subsection{Object Detectors with Conventional Images }
    In recent years, SSD\cite{liu2016ssd}, YOLO\cite{redmon2017yolo9000} and RetinaNet\cite{lin2017focal} have been proposed to achieve better performance in object detection. Most of one-stage methods have better inference speed, but their accuracy trails that of two-stage methods. However, recently RetinaNet\cite{lin2017focal} has achieved better performance on both speed and accuracy compared with state-of-the-art two stage models, Faster RCNN\cite{ren2015faster}, with a 4.0 \% improvement on COCO data\cite{lin2014microsoft}.  RetinaNet\cite{lin2017focal} combines the advantages of the SSD\cite{liu2016ssd} and YOLO\cite{redmon2017yolo9000} networks by performing a  multi-layer feature extraction and then feeding them into a box and class sub-network to generate final outputs. To avoid the class imbalance problem, RetinaNet\cite{lin2017focal} uses Focal Loss which is to address the one-stage detector problem in which there is an extreme imbalance between foreground and background classes during training.
    
	 However, many more popular detectors use a two-stage detection paradigm. The first model used is a R-CNN\cite{girshick2014rich}, which uses selective search to generate a sparse set of region proposals that may contain all objects while filtering out the majority of negative samples. In the second stage, a machine learning classifier is applied to classify the proposals into foreground and background. As R-CNNs\cite{girshick2014rich} become more and more popular, more advanced two-stage networks such as the Fast R-CNN\cite{girshick2015fast}, Faster R-CNN\cite{ren2015faster} and FPN\cite{lin2017feature} were developed. Different region proposal strategies such as RPN\cite{ren2015faster} and FPN\cite{lin2017feature} have also been used for region proposal generation. 
	\subsection{Object Detection in Aerial Images}
    Extensive work has been devoted to detecting objects in aerial images, drawing upon recent advances in Computer Vision and partially due to the  high demands for accurate aerial image applications. Most work tries to use object detectors that have achieved good performance in natural images and apply them into aerial images. However, due to the previously discussed properties of aerial images, some of this work can not be directly transferred. Thus, recently, researchers\cite{tang2017vehicle,sommer2017deep,yang2018automatic} have proposed different approaches based on fine-tuning networks pretrained on ImageNet\cite{deng2009imagenet} and COCO data\cite{lin2014microsoft} for detection in the aerial image domain. Since most objects in aerial images are quite small, such as small vehicles, so that researchers in computer vision for small object application majority focus on the aerial images with fine-tuning and modified object detector on natural images. 
    \subsection{Hard Example Mining}
    In general, mining hard positives enables the model to discover and expand sparsely sampled minority class boundaries, while mining hard-negatives aims to improve the margins of minority class boundaries corrupted by visually very similar imposter classes. In general, random sampling techniques are used to ameliorate the class imbalance problem \cite{provost2000machine}. Recent works demonstrate mining hard example would be able to improve the performance of imbalance classification problem \cite{dong2017class}. In addition, in two-stage detection, after generating multiple proposals from selective search using a R-CNN\cite{girshick2014rich}, the classifier needs to make a correct classification of all of the positive and negative proposals. Hard negative mining is necessary in this phase and results in good performance. Recently, Online Hard Example Mining (OHEM)\cite{shrivastava2016training} has been proposed which is an online bootstrapping algorithm for training region-based ConvNet object detectors like Fast R-CNN\cite{girshick2015fast}. OHEM simplifies training by removing some heuristics and hyper parameters and leads to better convergence. In the one-stage detector context, specifically SSD\cite{liu2016ssd}, the ratio of positive and negative with random sample which leads to a faster and more stable training.

        \begin{figure*}[h]
    \center
    \begin{subfigure}[b]{\linewidth}
    	\centerline{\includegraphics[trim=0 200 200 0, clip, width=0.6\textwidth,height=5cm]{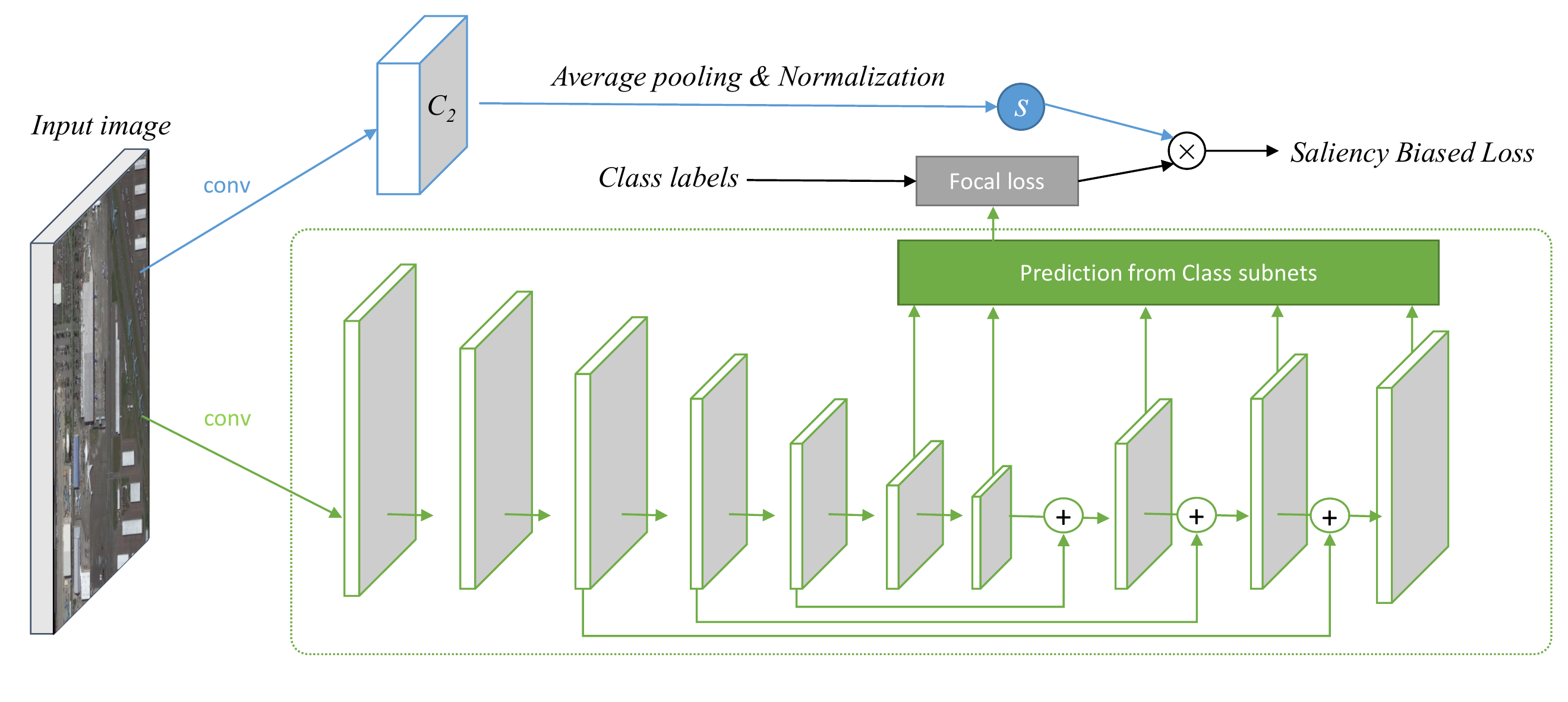}}
   	\end{subfigure}	
   	\caption{Salience Biased Loss. The bottom (green) network is RetinaNet. The top architecture (blue) is salience estimator network. In top path, C2 is the activation of conv2 of ResNet50. Salience Biased Loss will be generated by multiplying the Focal Loss of the bottom network with the average activation of the salience estimator.}
   	\end{figure*}	
\section{Salience Biased Loss}
    In this section, we propose a novel loss function, Salience Biased Loss, which can be used to improve performance of state-of-art object detection models with the same inference speed of model. Our novel loss focuses on training on a set of complicated images and prevents the vast number of easy cases from driving the converge of loss of the detector during training phase. The second part of this section will use RetinaNet to derive a deep learning network architecture, SBL-RetinaNet, which is used to evaluate the effectiveness of our loss function.

    \subsection{Salience Biased Loss}	
	In current implementations of object detection, the majority train on all provided images and then converge the multi-task loss function in order to locate and label the object. However, this kind of training strategy may impose potential problems. All input images are treated equally in the training phase so that the trained model may focus on the majority of easy cases. In the inference phase, due to variation of test images or the complexity of natural images in the real application, the model may fail to generalize and not perform accurately in many cases. To avoid this issue, hard example mining may be a good solution. Researchers focus on the difficulty of instance or number of objects for each objects to improve the performance of network. In this paper, we  use the salience of images to represent the complexity and propose a novel loss function, Salience Biased Loss, which will treat each image differently during the training phase. We introduce it beginning with the cross entropy and Focal Loss.
	\begin{equation}
		CE(p,y) =
		\begin{cases}
		-log(p), &if~y=1\cr  
		-log(1-p) &otherwise
		\end{cases}
		\label{eq:instance_prob}
    \end{equation}	
	where p is the probability output by the model for foreground, and y is ground-truth class. For convenience, we use CE(p,y)=-log(\(p_{t}\)).
	\begin{equation}
		p_{t} =
		\begin{cases}
		p, &if~y=1\cr  
		1-p &otherwise
		\end{cases}
		\label{eq:instance_prob}
    \end{equation}
    
    From the curve of cross-entropy, we notice that even very easy examples may incur a loss with non-trivial magnitude such that all these small losses my overwhelm the rare class loss when the number of easy samples is large. In order to avoid the class-imbalance and easy/hard example problem in the middle one-stage detector, recent work has uses Focal Loss to improve the performance of classification of foreground/background. Focal Loss can prevent the easily-classified negatives from overwhelming the majority of the loss and dominating the gradient. The idea is pretty straight-forward: introduce a weight factor \(\alpha\) for foreground and 1-\(\alpha\) for background to avoid an imbalance. In addition, adding another module factor \(\left ( 1-p_{t} \right )^{\gamma }\) to cross-entropy loss with tuning parameter \(\gamma\) is necessary. The reason is that if the example is mis-classified, the module factor will be near 1 and the loss function will remain constant; however, if the example is in the correct class, the module will scale the weight near to 0 so that the importance of the easy class in the loss function will be dropped diminished. With the tuning parameter \(\gamma\), the scale of importance of the module factor can be tuned based on the experiment. Recent work suggests that \(\gamma\)=2 and \(\alpha\)=0.25 results in the best performance in natural image object detection. The loss function then becomes:
	\begin{equation}
		Loss(p,y)=\alpha*\left ( 1-p_{t} \right )^{\gamma }*CE(p,y)
    \end{equation}	
    where \(\gamma\)=2 and \(\alpha\)=0.25 in our experiments.
    
    After addressing the imbalance and easy/hard problem in the middle of the object detector, we consider how to scale the difficulty of input images in order to prevent simple cases from overwhelming loss converge. We define simple cases as those whose images with a simple background and little noise information, and hard cases as those which are more complicated than simple cases. Detecting the difficulty of images is very challenging without human labels. Some existing works use Edgebox\cite{zitnick2014edge} to determine the complexity of images, like WiderFace\cite{yang2016wider}. This may not be an optimial solution because it requires too much computation and is hard to decide which is the best threshold. In the DOTA data, we use a pretrained deep learning architecture to decide the complexity of images. Our idea is that there will be more noisy information if there are more activated neural in reception field of each layers. The state-of-the-art deep learning architecture has been trained by large-scale data, ImageNet\cite{deng2009imagenet}, and the capability of network can learn the features of normal object in the images. Based on this condition, we use a network pretrained by ImageNet\cite{deng2009imagenet} as a salience estimator and then extract features from different convolution layers to represent the complexity of images. This yields a representation of complexity of images such that:
	\begin{equation}
		S=\frac{1}{C*W*H}\sum_{c=1}^{C}\sum_{w=1}^{W}\sum_{h=1}^{H}f_{c,w,h}(x)
    \end{equation}	
    where \(S\) is the average activation indicative of salience level of input images, x is the input image, \(f_{c,w,h}\) is a vonvolution block with output of dimension W*H*C. Based on this formula, the easy case has less activated neurons and will result in a smaller value. Each image may have a varied representation of complexity due to the different scale of convolution blocks in deep learning architectures. We consider all of them together in our experiments since it captures a multi-scale view for each image and we will see which part of scale have the biggest affection on single image. 
    
    Once we consider the complexity of each image, one common way to use this information is to augment training samples by up-sampling if the sample is hard cases. However, this way will give more training burden and the loss is hard to converge into global minimum if the data is augmented too heavily. In our implementation, we put complexity information of images into a loss function instead of focusing on doing data augmentation. In the experiments, we set the complexity of images as a weight function and include it in the classification loss function. We call it Salience Biased Loss since the loss is based on the salience information of each image and all the training images are biased during the training phase. Our final loss formula is below:
	\begin{equation}
		SBL(p,y)=S*Loss(p,y)
    \end{equation}	
    where SBL stands for Salience Biased Loss. The loss of easy cases will become smaller when the S value is small. We note two properties of the Salience Biased Loss. (1) When the loss converges, the hard cases will contribute more, and the easy cases will contribute less. Because the easy case will have very small loss value, the loss will be down-weighted. (2) Since S will be extracted from different convolution blocks, multi-scale feature information will be included in the loss function. For instance, the lower level features have larger feature map, and each point in its feature map represents a small object in the original images. If the value is higher, the noisy small object information will be put into the loss function.

    \subsection{SBL-RetinaNet}
	In order to make Salience Biased Loss work, we use RetinaNet\cite{lin2017focal}, which is the state-of-art one-stage detector to derive a network, called SBL-RetinaNet. As shown in Figure 2, the main architecture is the same as RetinaNet since our purpose is to keep the original network structure with respect to inference time while using a different strategy to improve performance. To make improve performance for aerial images, some modification of models is still necessary. We will present modifications made in a later session. Compared with standard RetinaNet, SBL-RetinaNet adds one more pretrained network and fixes all weights in the convolution block to extract features from different convolution layers. In our experiment, ResNet50\cite{he2016deep} is used to test the performance of the model, same as the backbone in RetinaNet\cite{lin2017focal}. In terms of feature extraction, the output of conv2, conv3, conv4, conv5 of ResNet50 will be denoted as \{C2,C3,C4,C5\} for convenience. These features are used to estimate salience complexity of input images in our experiment. Each extracted feature will be used as a weight into loss function to make training sample into different scenario in training.

\section{Experiments}	
    We present experimental results on the largest benchmark dataset of aerial images, DOTA, and small object dataset of aerial images, LIBAI. For all the ablation study will be tested in DOTA dataset. Experimental results  will be presented based on both datasets.
    
	\subsection{Dataset}
	DOTA is the largest and most diverse published dataset for multi-class object detection in aerial images. There are 2806 images in total collected from variety of camera sensors. The images are mainly acquired from Google Earth and the rest of them are from China Center for Resources Satellite Data and Application. In this large scale dataset, 15 categories need to be classified: plane, baseball diamond (BD), bridge, ground field track (GTF), small vehicle (SV), large vehicle (LV), tennis court (TC), basketball court (BC), storage tank (SC), soccer ball field (SBF), roundabout (RA), swimming pool (SP), helicopter (HC), and harbor. Across all these categories, 57\% images are small objects which are within 50 *50 pixels and others occupy different scales, making the task more difficult than conventional object detection scenarios. The DOTA dataset is split into training(1/2), validation(1/6), and test (1/3) sets.
    
    The LBAI dataset was provided by the Missouri Department of Conservation (MDC). The images are taken for studies to protect the living habitats of birds. The total dataset has 230 GB of data with 440 high-resolution images of 5760 pixels by 3840 pixels. 
    Many of the objects in LBAI are small (less than 40*40) with a variety of colors, scales, shapes and poses of the objects, as well as resolution of the background. Since LBAI only focuses on the bird detection, there is only one category needs to be detected. 
    
    \subsection{Evaluation}
    In order to assess the accuracy of the detection method, mean Average Precision (mAP) as for PASCAL VOC\cite{everingham2010pascal} is provided for the DOTA dataset. For the final results, we follow the original work in DOTA paper in order to make a comparison with other state-of-art detectors. Based on our experiments, we found different NMS strategies, like Soft-NMS\cite{bodla2017soft}, are able to improve the test performance, however, it may not be fair to compare with the original results provided. In our experiment, we focus on the HBB task in DOTA and set non-maximum suppression (NMS) as 0.3 for all categories after generating results based on the predicted class. IoU ratio of predicted and ground truth boxes are using 0.5, as fashion in the object detection domain.
    
    In terms of LBAI data, we also focus on the official way in order to make a comparison with other state-of-art results. In the LBAI paper, they use F1 score as final metrics, with the NMS parameter also 0.1 since the object is too small. In our experiment, we focus on LBAI-A dataset since that is the most typical case for aerial image applications. 

    \subsection{RetinaNet Modification}	
    To make RetinaNet work with aerial images, modification was necessary. In our final results, we also provide the results of the original RetinaNet without any change to make comparison. In terms of modification, without influence of inference time, we only change aspect ratios to {1:3,1:1,3:1}, and anchor sizes {\(2^{1}\),\(2^{1/2}\),\(0.3\)}. In our experiments, other modifications are also tested, like modification of anchor numbers to 15 or 20 with a higher aspect ratio and anchor sizes, however, the results are poor. The reason for extending the ratio of one edge of the aspect ratio is that there are special categories such as bridge or harbor which is a long rectangle in the DOTA dataset. We use a 0.3 anchor size in order to detect small objects in aerial images and to consider different scales of objects.
    
    \subsection{Ablation Study}
    \subsubsection{Image Complexity Analysis}
    In our experiment, we use a deep neural network to define the complexity of images. When the background of an image has more noisy information, the image will be more complex. In order to prove that this assumption is correct, we use a heuristic method to demonstrate different backgrounds of DOTA, as shown in Figure 4. In Figure 4, it shows top 10 largest and smallest values of salience information based on multi-scale of feature maps of deep neural network. From the figure, it demonstrates all top images are more noisy than bottom ones and shows variation of scale of noisy information to describe global information of images.
        \begin{table}[htbp]
    	\caption{Evaluation of \(new\_\min\) in Salience Normalization}
    	\begin{center}
    	\begin{tabular}{|c|c|c|}
    	\hline

    	\textbf{\(new\_\min\)}&\textbf{Normalization}&\textbf{mAP} \\ \hline
    	0.3 & Y & 61.79 \\
    	0.5 & Y & \textbf{63.48} \\
    	0.7 & Y & 62.4 \\
    	1 & Y & 62.51 \\ 
    	- & N & 62.86 \\\hline
    	\end{tabular}
    	\label{tab:final}
    	\end{center}
    \end{table}
    
    However, once we add this complexity information into loss function, the gains in improvement are what we care about. In our experiments, we use the modified RetinaNet\cite{lin2017focal} as a baseline, and then extracted features such as complexity information from the last residual stage C5 of ResNet50\cite{he2016deep}. To make a fair comparison, all other parameters are held the same. As we see from Table III, there is a 0.35 mAP improvement from 62.51 to 62.86. When we run experiments multiple times, the stand deviation for both models is 0.017 and 0.022, respectively which means the 0.35 AP improvement is a significant difference between new models and the modified RetinaNet\cite{lin2017focal}.

        \begin{figure}[h!]
  	\centering
   	\begin{subfigure}[b]{0.4\linewidth}
    	\includegraphics[width=\linewidth]{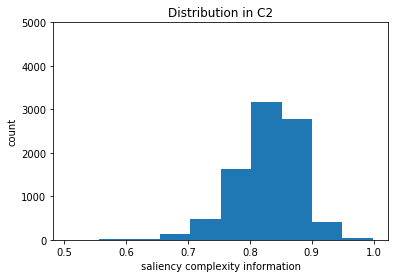}
   	\end{subfigure}
  	\begin{subfigure}[b]{0.4\linewidth}
    	\includegraphics[width=\linewidth]{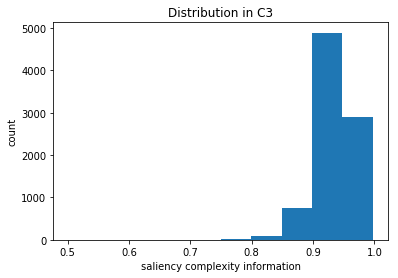}
  	\end{subfigure}
  	   	\begin{subfigure}[b]{0.4\linewidth}
    	\includegraphics[width=\linewidth]{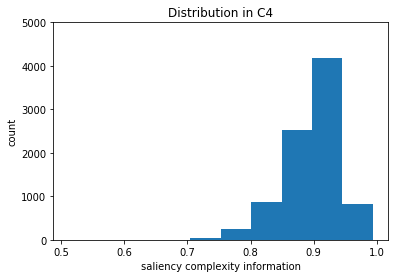}
   	\end{subfigure}
  	\begin{subfigure}[b]{0.4\linewidth}
    	\includegraphics[width=\linewidth]{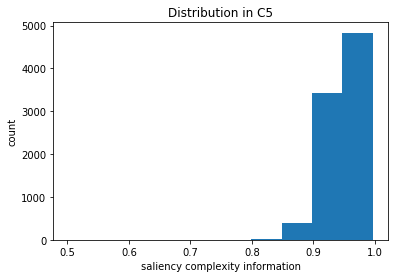}
  	\end{subfigure}
  	\caption{Distribution of Salience information for each residual block of ResNet50}
  	\label{fig:sample data}
	\end{figure}

	\begin{figure*}[h]
   	\center
    \begin{subfigure}[b]{\linewidth}
    	\includegraphics[width=\textwidth,height=2cm]{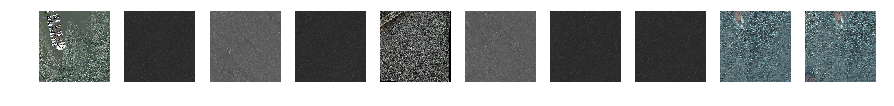}
   	\end{subfigure}\\[-2ex]
   	\begin{subfigure}[b]{\linewidth}
      \includegraphics[width=\textwidth,height=2cm]{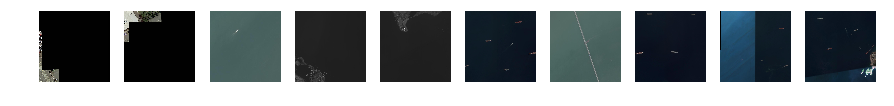}
      \abovecaptionskip-\belowcaptionskip
      \caption{}
    \end{subfigure}
   	\begin{subfigure}[b]{\linewidth}
      \includegraphics[width=\textwidth,height=2cm]{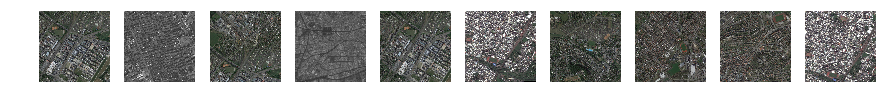}
    \end{subfigure}\\[-2ex]
   	\begin{subfigure}[b]{\linewidth}
      \includegraphics[width=\textwidth,height=2cm]{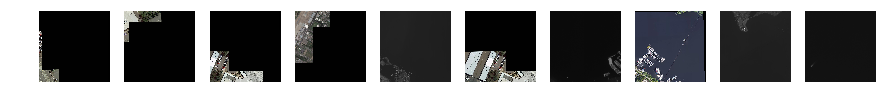}
      \abovecaptionskip-\belowcaptionskip
      \caption{}
    \end{subfigure}
   	\begin{subfigure}[b]{\linewidth}
      \includegraphics[width=\textwidth,height=2cm]{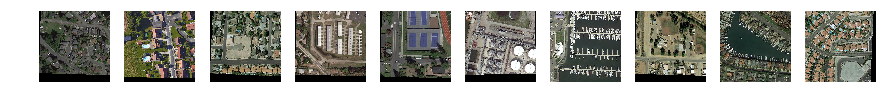}
    \end{subfigure}\\[-2ex]
   	\begin{subfigure}[b]{\linewidth}
      \includegraphics[width=\textwidth,height=2cm]{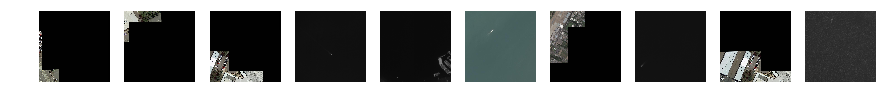}
      \abovecaptionskip-\belowcaptionskip
      \caption{}
    \end{subfigure}
   	\begin{subfigure}[b]{\linewidth}
      \includegraphics[width=\textwidth,height=2cm]{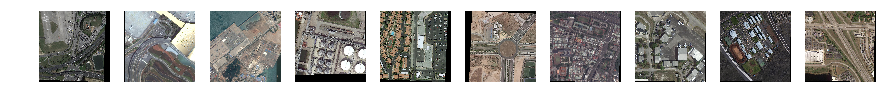}
    \end{subfigure}\\[-2ex]
   	\begin{subfigure}[b]{\linewidth}
      \includegraphics[width=\textwidth,height=2cm]{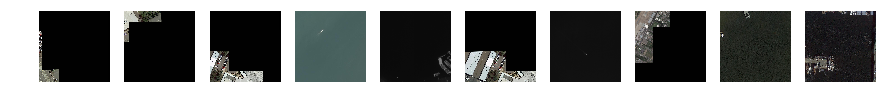}
      \caption{}
    \end{subfigure}
    \caption{Multi-scale Salience Analysis. Top images in (a),(b),(c),(d) are the top 10 largest values from C2-C5 of ResNet50. Bottoms in (a),(b),(c),(d) are top 10 smallest values. Each pairs of images in (a),(b),(c),(d) demonstrate the salience level for the input data. All top images are more noisy than bottom ones. Compare across all top images in (a),(b),(c),(d), which shows variation of scale of noisy information and provides a different angle of description of noise for images.}
    \end{figure*}
    \subsubsection{Salience Normalization}
     We look into the distribution of salience level for each convolution block. As shown in Figure 3, the difference of salience level of images is fairly minimal such that no significant information can be used between easy and hard cases. To address this issue, we apply normalization to extend the difference between easy and hard cases. The formula is below:
	\begin{equation}
		S'=\frac{S-min}{max-min}(new\_\max-new\_\min)+new\_\min
    \end{equation}	
    where S is the original salience level, min and max are calculated by training data before the training phase, \(new\_\max\) is set as 1, \(new\_\min\) are in the list of (0.3,0.5,0.7). In our experiments, we test the performance of \(new\_\min\) with different values on the C5 residual block of ResNet50\cite{he2016deep}. Table I shows \(new\_\min\) set as 0.5 giving the best performance with 0.62 mAP improvement compared with the DOTA dataset.

    \begin{table}[htbp]
    	\caption{Evaluation of multi-scale features in Salience Biased Loss}
    	\begin{center}
    	\begin{tabular}{|c|c|c|c|c|}
    	\hline

    	\textbf{Conv Block}&\textbf{C2}&\textbf{C3}&\textbf{C4}&\textbf{C5} \\ \hline
    	mAP &\textbf{64.77} & 64.51& 63.32& 63.48\\ \hline
    	\end{tabular}
    	\label{tab:final}
    	\end{center}
    \end{table}
    \subsubsection{Multi-scale Salience Analysis}
    To understand which feature layers provides more useful information to our loss function, multi-scale salience features are tested on the DOTA dataset. In our experiments, we estimate salience level using C2-C5 from ResNet50\cite{he2016deep} with normalization. For each block's noisy information, Salience Biased Loss will be generated and then fed to RetinaNet\cite{lin2017focal}. In our experiments, compared with modified RetinaNet\cite{lin2017focal}, the salience level generated from C2 provides the best results with 2.26 mAP improvement compared with modified RetinaNet, as shown in the Table II.

	\begin{table*}[t]
	  \caption{Results of DOTA test dataset}
      \centering
      \begin{tabular}{cccccccc}
      \hline

\textbf{Models} & \textbf{YOLO\cite{redmon2017yolo9000}} & \textbf{SSD\cite{liu2016ssd}} & \textbf{RFCN\cite{dai2016r}} & \textbf{FR-H\cite{ren2015faster}}  & \textbf{RetinaNet\cite{lin2017focal}} & \textbf{RetinaNet*} & \textbf{SBL-RetinaNet} \\ \hline
\textbf{Plane}  & 76.9          & 44.74        & 79.33         & 80.32          & 78.22              & 89.03               & \textbf{89.15}         \\
\textbf{BD}     & 33.87         & 11.21        & 44.26         & 77.55          & 53.41              & 62.14               & \textbf{66.04}         \\
\textbf{Bridge} & 22.73         & 6.22         & 36.58         & 32.86          & 26.38              & 43.88               & \textbf{46.79}         \\
\textbf{GTF}    & 34.88         & 6.91         & 53.53         & \textbf{68.13} & 42.27              & 47.05               & 52.56                  \\
\textbf{SV}     & 38.73         & 2            & 39.38         & 53.66          & 63.64              & \textbf{73.57}      & 73.06                  \\
\textbf{LV}     & 32.02         & 10.24        & 34.15         & 52.49          & 52.63              & 65.18               & \textbf{66.13}         \\
\textbf{Ship}   & 52.37         & 11.34        & 47.29         & 50.04          & 73.19              & 78.65               & \textbf{78.66}         \\
\textbf{TC}     & 61.65         & 15.59        & 45.66         & 90.41          & 87.17              & \textbf{90.86}      & 90.85                  \\
\textbf{BC}     & 48.54         & 12.56        & 47.74         & \textbf{75.05} & 44.64              & 66.28               & 67.4                   \\
\textbf{ST}     & 33.91         & 17.94        & 65.84         & 59.59          & 57.99              & 70.26               & \textbf{72.22}         \\
\textbf{SBF}    & 29.27         & 14.73        & 37.92         & \textbf{57}    & 18.03              & 35.07               & 39.88                  \\
\textbf{RA}     & 36.83         & 4.55         & 44.23         & 49.81          & 51                 & \textbf{58.26}      & 56.89                  \\
\textbf{Harbor} & 36.44         & 4.55         & 47.23         & 61.69          & 43.39              & 68.93               & \textbf{69.58}         \\
\textbf{SP}     & 38.26         & 0.53         & 50.64         & 56.46          & 56.56              & 66.34               & \textbf{67.73}         \\
\textbf{HC}     & 11.61         & 1.01         & 34.9          & \textbf{41.85} & 7.44               & 22.16               & 34.74                  \\ \hline
\textbf{mAP}    & 39.2          & 10.94        & 47.24         & 60.46          & 50.39              & 62.51               & \textbf{64.77}        \\ \hline
        \multicolumn{4}{l}{$^*$ RetinaNet with modified anchor sizes and ratios}
      \end{tabular}
      \label{tab:final}
    \end{table*}    
    \subsection{Training Detector}
    During our training phase, we run ResNet50\cite{he2016deep} as a backbone of RetinaNet\cite{lin2017focal} with the same architecture as the noisy information extraction part. The inputs are 1024*1024 for DOTA and 512*512 for LIBAI. We use Adam as the optimization methodology. Two 1080Ti has been used in our experiments with batch size as 2. Unless otherwise specified, all models are trained for 200k iterations with an initial learning rate of 0.0001, which is then divided by 10 after each 80k iterations. The training loss is the sum of Salience Biased Loss and the standard smooth L1 loss used for classification and box regression, respectively.
    \begin{figure}[h!]
  	\centering
   	\begin{subfigure}[b]{0.4\linewidth}
    	\includegraphics[width=\linewidth]{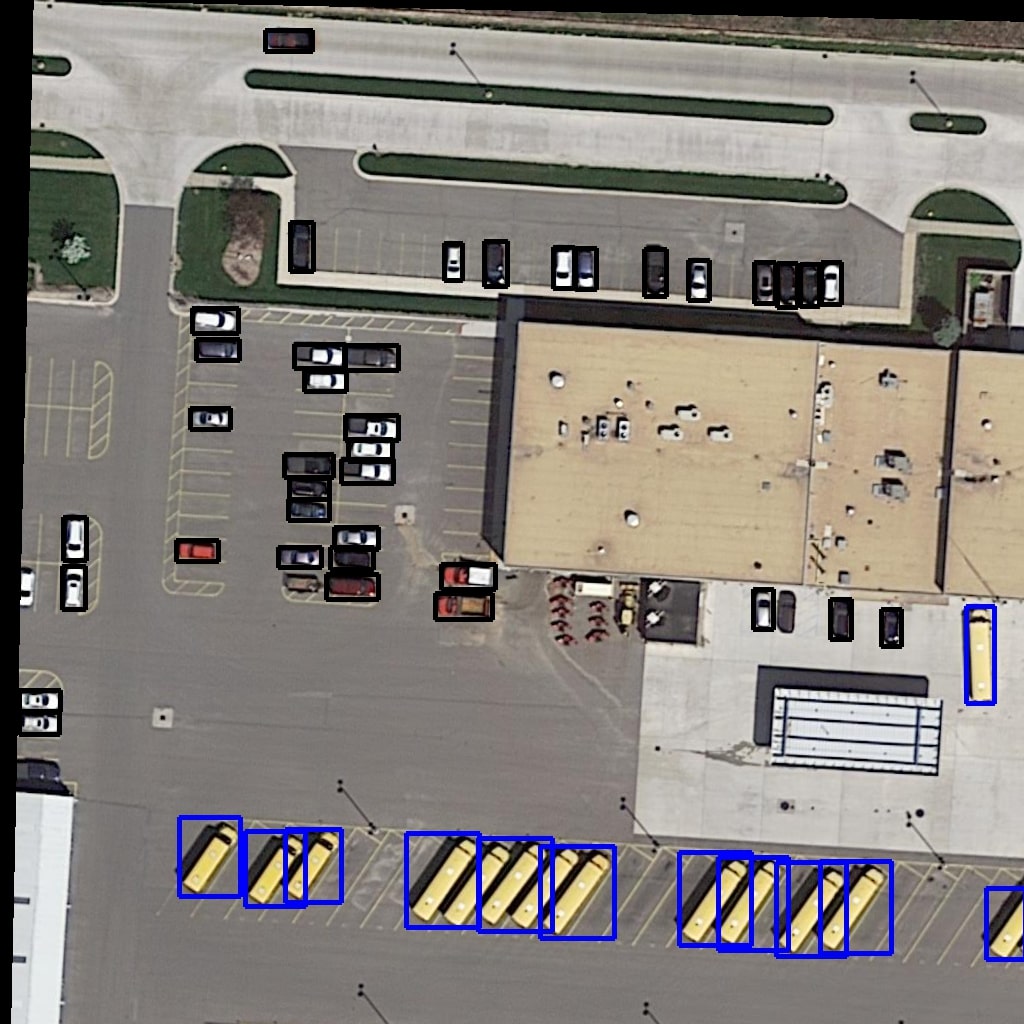}
   	\end{subfigure}
  	\begin{subfigure}[b]{0.4\linewidth}
    	\includegraphics[width=\linewidth]{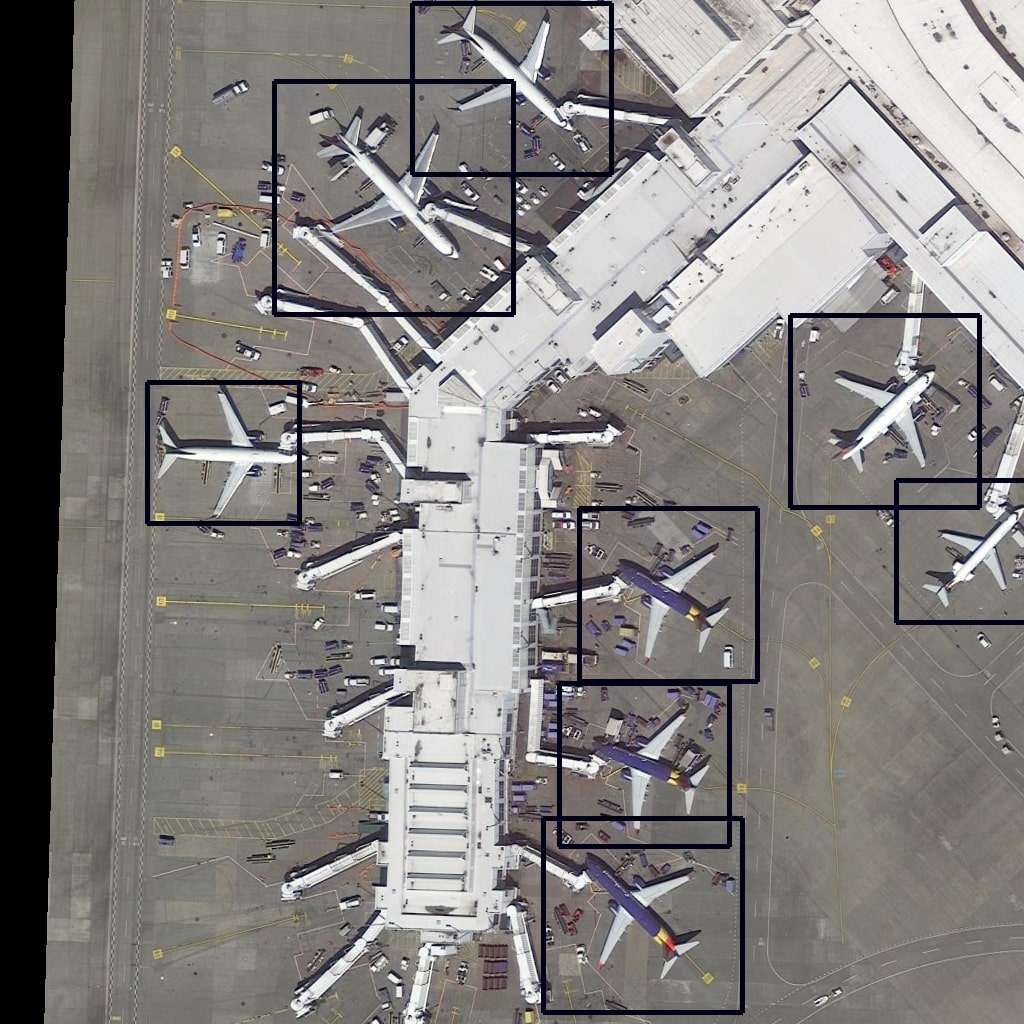}
  	\end{subfigure}
  	   	\begin{subfigure}[b]{0.4\linewidth}
    	\includegraphics[width=\linewidth]{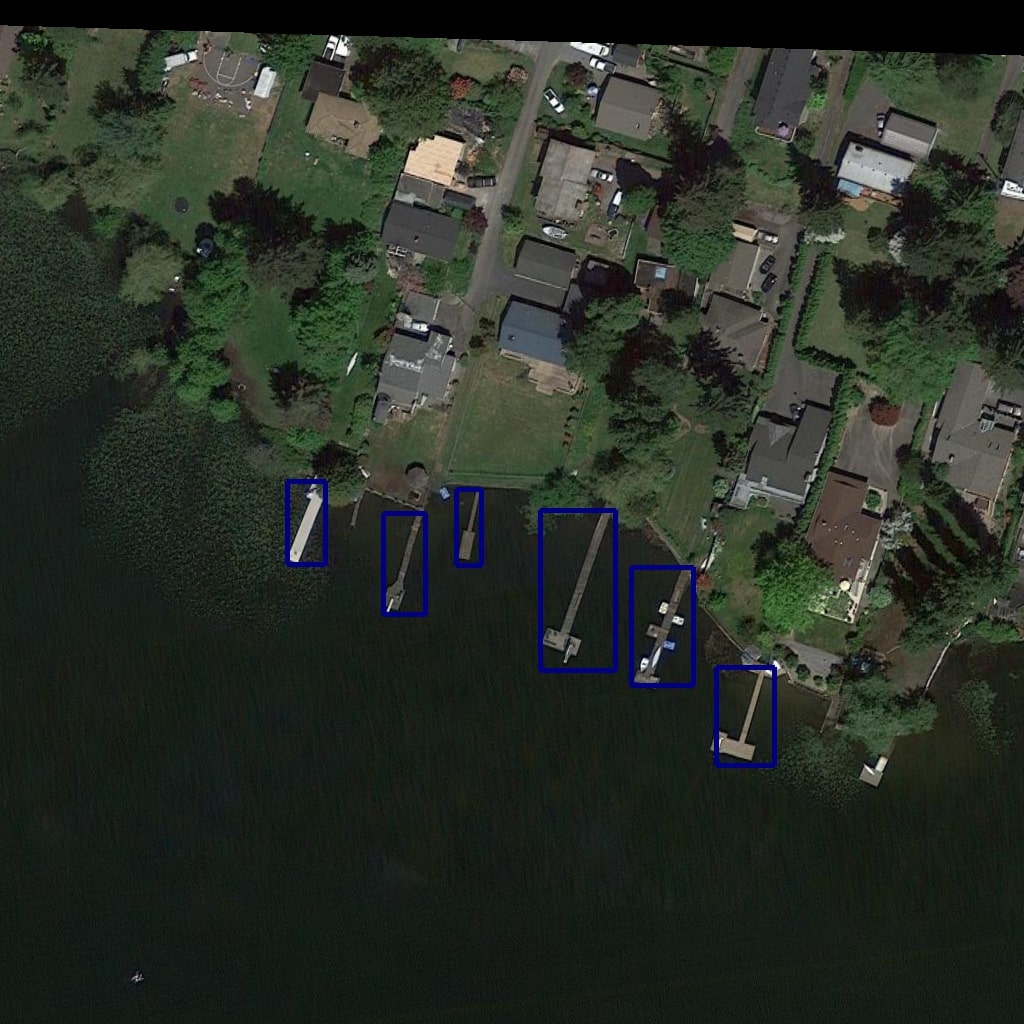}
   	\end{subfigure}
  	\begin{subfigure}[b]{0.4\linewidth}
    	\includegraphics[width=\linewidth]{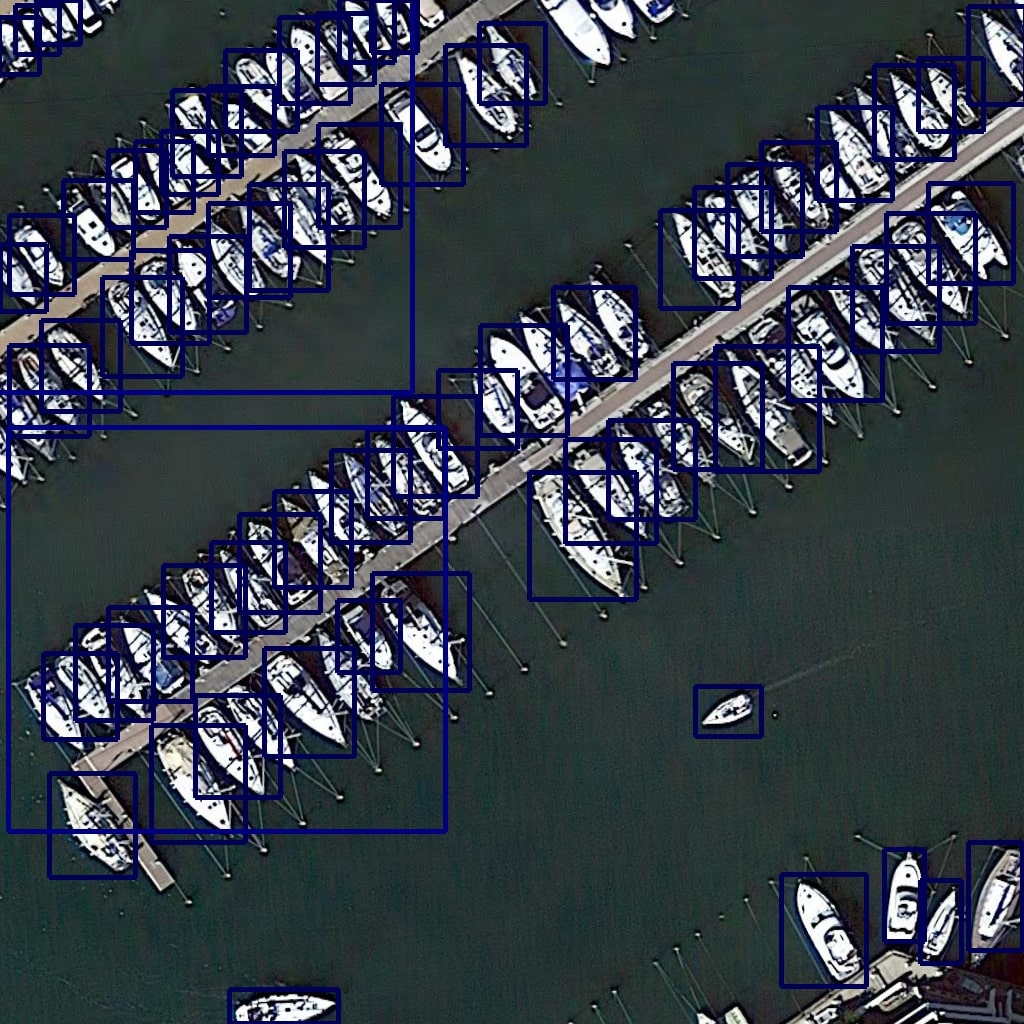}
  	\end{subfigure}
  	\caption{Sample output of  images of test data in DOTA (HBB)}
  	\label{fig:sample data}
	\end{figure}
    \subsection{Experimental Results on DOTA}
    We evaluate our SBL-RetinaNet with Salience Biased Loss on the largest aerial image dataset, DOTA and compare the performance on test data to recent state-of-art methods including both one-stage and two-stage models. Result in Table III demonstrate our proposed method achieves 4.31 mAP improvement with the closest competitor, Faster R-CNN\cite{ren2015faster}. Figure 5 demonstrates the example outputs from the HBB (Horizontal Bounding Box) in the DOTA dataset. Compared with  our modified models of RetinaNet\cite{lin2017focal}, our Salience Biased Loss can help to improve 2.26 mAP as compared with not changing the original network architecture. Considered with the inference running speed, without adding any more parameter in network, the speed are the same as original RetinaNet\cite{lin2017focal}. 
   
    \subsection{Generalization on LIBAI}
    As shown in Table 4, our algorithm also significantly improves upon the baseline in LIBAI-A. Our proposed method is better than other current state-of-art networks with a 2.73\% improvement on the F1 score, and also a 1.31\% improvement for the modified RetinaNet\cite{lin2017focal}. This suggests that our proposed method has generalizes and generates competitive results on the aerial images dataset.
 \begin{table}[htbp]
    	\caption{Results of LIBAI-A dataset}
    	\begin{center}
    	\begin{tabular}{|c|c|c|c|c|}
    	\hline

    	\textbf{Models}&\textbf{Precision}&\textbf{Recall}&\textbf{F1} \\ \hline
    	 YOLO v3\cite{redmon2018yolov3} &88.69\% &90.85\%&89.76\%\\
    	 SSD\cite{liu2016ssd} & 20.23\% &86.9\% &32.82\%\\
         Mask R-CNN\cite{he2017mask} &77.2\% &84.2\% &80.5\% \\

    	 RetinaNet\cite{lin2017focal} &88.87\% &89.81\% &89.34\% \\
         RetinaNet* &90.64\% &91.72\% &91.18\% \\
         SBL-RetinaNet &\textbf{90.74}\% &\textbf{94.31}\% &\textbf{92.49}\% \\ \hline 
         \multicolumn{4}{l}{$^*$ RetinaNet with modified anchor sizes and ratios}
    	\end{tabular}
    	\label{tab:final}
    	\end{center}
    \end{table}
\section{Conclusion}
    In this paper, we proposed a novel loss function, Salience Biased Loss, and developed a variant of RetinaNet called SBL-RetinaNet. SBL-RetinaNet outperformed existing  state-of-art deep learning object detectors, significantly, with at least 4.31 mAP improvement on the largest aerial image object detection dataset, DOTA. The goal of this work is to develop a new training strategy to actively weight training images based on image complexities in order to improve the performance of deep learning architectures without changing inference speed. This paper proposed a new way to represent image complexity as the salience information extracted from convolution blocks of a pre-trained deep learning model. In the future, salience level ensembles and local salience information will be investigated further for aerial image object detection.

\bibliographystyle{./IEEEtran}
\bibliography{./IEEEabrv,./IEEEexample}

\begin{thebibliography}{10}
\providecommand{\url}[1]{#1}
\csname url@samestyle\endcsname
\providecommand{\newblock}{\relax}
\providecommand{\bibinfo}[2]{#2}
\providecommand{\BIBentrySTDinterwordspacing}{\spaceskip=0pt\relax}
\providecommand{\BIBentryALTinterwordstretchfactor}{4}
\providecommand{\BIBentryALTinterwordspacing}{\spaceskip=\fontdimen2\font plus
\BIBentryALTinterwordstretchfactor\fontdimen3\font minus
  \fontdimen4\font\relax}
\providecommand{\BIBforeignlanguage}[2]{{%
\expandafter\ifx\csname l@#1\endcsname\relax
\typeout{** WARNING: IEEEtran.bst: No hyphenation pattern has been}%
\typeout{** loaded for the language `#1'. Using the pattern for}%
\typeout{** the default language instead.}%
\else
\language=\csname l@#1\endcsname
\fi
#2}}
\providecommand{\BIBdecl}{\relax}
\BIBdecl

\bibitem{krizhevsky2012imagenet}
A.~Krizhevsky, I.~Sutskever, and G.~E. Hinton, ``Imagenet classification with
  deep convolutional neural networks,'' in \emph{Advances in neural information
  processing systems}, 2012, pp. 1097--1105.

\bibitem{simonyan2014very}
K.~Simonyan and A.~Zisserman, ``Very deep convolutional networks for
  large-scale image recognition,'' \emph{arXiv preprint arXiv:1409.1556}, 2014.

\bibitem{he2016deep}
K.~He, X.~Zhang, S.~Ren, and J.~Sun, ``Deep residual learning for image
  recognition,'' in \emph{Proceedings of the IEEE conference on computer vision
  and pattern recognition}, 2016, pp. 770--778.

\bibitem{girshick2014rich}
R.~Girshick, J.~Donahue, T.~Darrell, and J.~Malik, ``Rich feature hierarchies
  for accurate object detection and semantic segmentation,'' in
  \emph{Proceedings of the IEEE conference on computer vision and pattern
  recognition}, 2014, pp. 580--587.

\bibitem{ren2015faster}
S.~Ren, K.~He, R.~Girshick, and J.~Sun, ``Faster r-cnn: Towards real-time
  object detection with region proposal networks,'' in \emph{Advances in neural
  information processing systems}, 2015, pp. 91--99.

\bibitem{he2017mask}
K.~He, G.~Gkioxari, P.~Doll{\'a}r, and R.~Girshick, ``Mask r-cnn,'' in
  \emph{Computer Vision (ICCV), 2017 IEEE International Conference on}.\hskip
  1em plus 0.5em minus 0.4em\relax IEEE, 2017, pp. 2980--2988.

\bibitem{lin2017feature}
T.-Y. Lin, P.~Doll{\'a}r, R.~B. Girshick, K.~He, B.~Hariharan, and S.~J.
  Belongie, ``Feature pyramid networks for object detection.'' in \emph{CVPR},
  vol.~1, no.~2, 2017, p.~3.

\bibitem{redmon2017yolo9000}
J.~Redmon and A.~Farhadi, ``Yolo9000: better, faster, stronger,'' \emph{arXiv
  preprint}, 2017.

\bibitem{dai2016r}
J.~Dai, Y.~Li, K.~He, and J.~Sun, ``R-fcn: Object detection via region-based
  fully convolutional networks,'' in \emph{Advances in neural information
  processing systems}, 2016, pp. 379--387.

\bibitem{shrivastava2016training}
A.~Shrivastava, A.~Gupta, and R.~Girshick, ``Training region-based object
  detectors with online hard example mining,'' in \emph{Proceedings of the IEEE
  Conference on Computer Vision and Pattern Recognition}, 2016, pp. 761--769.

\bibitem{liu2016ssd}
W.~Liu, D.~Anguelov, D.~Erhan, C.~Szegedy, S.~Reed, C.-Y. Fu, and A.~C. Berg,
  ``Ssd: Single shot multibox detector,'' in \emph{European conference on
  computer vision}.\hskip 1em plus 0.5em minus 0.4em\relax Springer, 2016, pp.
  21--37.

\bibitem{chen2017automatic}
G.~Chen, P.~Sun, and Y.~Shang, ``Automatic fish classification system using
  deep learning,'' in \emph{Tools with Artificial Intelligence (ICTAI), 2017
  IEEE 29th International Conference on}.\hskip 1em plus 0.5em minus
  0.4em\relax IEEE, 2017, pp. 24--29.

\bibitem{tang2017vehicle}
T.~Tang, S.~Zhou, Z.~Deng, H.~Zou, and L.~Lei, ``Vehicle detection in aerial
  images based on region convolutional neural networks and hard negative
  example mining,'' \emph{Sensors}, vol.~17, no.~2, p. 336, 2017.

\bibitem{sommer2017deep}
L.~W. Sommer, T.~Schuchert, and J.~Beyerer, ``Deep learning based
  multi-category object detection in aerial images,'' in \emph{Automatic Target
  Recognition XXVII}, vol. 10202.\hskip 1em plus 0.5em minus 0.4em\relax
  International Society for Optics and Photonics, 2017, p. 1020209.

\bibitem{yang2018automatic}
X.~Yang, H.~Sun, K.~Fu, J.~Yang, X.~Sun, M.~Yan, and Z.~Guo, ``Automatic ship
  detection in remote sensing images from google earth of complex scenes based
  on multiscale rotation dense feature pyramid networks,'' \emph{Remote
  Sensing}, vol.~10, no.~1, p. 132, 2018.

\bibitem{deng2009imagenet}
J.~Deng, W.~Dong, R.~Socher, L.-J. Li, K.~Li, and L.~Fei-Fei, ``Imagenet: A
  large-scale hierarchical image database,'' in \emph{Computer Vision and
  Pattern Recognition, 2009. CVPR 2009. IEEE Conference on}.\hskip 1em plus
  0.5em minus 0.4em\relax Ieee, 2009, pp. 248--255.

\bibitem{lin2014microsoft}
T.-Y. Lin, M.~Maire, S.~Belongie, J.~Hays, P.~Perona, D.~Ramanan,
  P.~Doll{\'a}r, and C.~L. Zitnick, ``Microsoft coco: Common objects in
  context,'' in \emph{European conference on computer vision}.\hskip 1em plus
  0.5em minus 0.4em\relax Springer, 2014, pp. 740--755.

\bibitem{lin2017focal}
T.-Y. Lin, P.~Goyal, R.~Girshick, K.~He, and P.~Doll{\'a}r, ``Focal loss for
  dense object detection,'' \emph{arXiv preprint arXiv:1708.02002}, 2017.

\bibitem{sun2016ada}
P.~Sun, N.~M. Wergeles, C.~Zhang, L.~M. Guerdan, T.~Trull, and Y.~Shang,
  ``Ada-automatic detection of alcohol usage for mobile ambulatory
  assessment,'' in \emph{Smart Computing (SMARTCOMP), 2016 IEEE International
  Conference on}.\hskip 1em plus 0.5em minus 0.4em\relax IEEE, 2016, pp. 1--5.

\bibitem{bernstein2017using}
J.~P. Bernstein, B.~J. Mendez, P.~Sun, Y.~Liu, and Y.~Shang, ``Using deep
  learning for alcohol consumption recognition,'' in \emph{Consumer
  Communications \& Networking Conference (CCNC), 2017 14th IEEE Annual}.\hskip
  1em plus 0.5em minus 0.4em\relax IEEE, 2017, pp. 1020--1021.

\bibitem{xia2018dota}
G.-S. Xia, X.~Bai, J.~Ding, Z.~Zhu, S.~Belongie, J.~Luo, M.~Datcu, M.~Pelillo,
  and L.~Zhang, ``Dota: A large-scale dataset for object detection in aerial
  images,'' in \emph{Proc. CVPR}, 2018.

\bibitem{li2018multiscale}
S.~Li, Z.~Zhang, B.~Li, and C.~Li, ``Multiscale rotated bounding box-based deep
  learning method for detecting ship targets in remote sensing images,''
  \emph{Sensors}, vol.~18, no.~8, p. 2702, 2018.

\bibitem{azimi2018towards}
S.~M. Azimi, E.~Vig, R.~Bahmanyar, M.~K{\"o}rner, and P.~Reinartz, ``Towards
  multi-class object detection in unconstrained remote sensing imagery,''
  \emph{arXiv preprint arXiv:1807.02700}, 2018.

\bibitem{liu2018performance}
Y.~Liu, P.~Sun, M.~R. Highsmith, N.~M. Wergeles, J.~Sartwell, A.~Raedeke,
  M.~Mitchell, H.~Hagy, A.~D. Gilbert, B.~Lubinski \emph{et~al.}, ``Performance
  comparison of deep learning techniques for recognizing birds in aerial
  images,'' in \emph{2018 IEEE Third International Conference on Data Science
  in Cyberspace (DSC)}.\hskip 1em plus 0.5em minus 0.4em\relax IEEE, 2018, pp.
  317--324.

\bibitem{girshick2015fast}
R.~Girshick, ``Fast r-cnn,'' in \emph{Proceedings of the IEEE international
  conference on computer vision}, 2015, pp. 1440--1448.

\bibitem{provost2000machine}
F.~Provost, ``Machine learning from imbalanced data sets 101.''

\bibitem{dong2017class}
Q.~Dong, S.~Gong, and X.~Zhu, ``Class rectification hard mining for imbalanced
  deep learning,'' 2017.

\bibitem{zitnick2014edge}
C.~L. Zitnick and P.~Doll{\'a}r, ``Edge boxes: Locating object proposals from
  edges,'' in \emph{European conference on computer vision}.\hskip 1em plus
  0.5em minus 0.4em\relax Springer, 2014, pp. 391--405.

\bibitem{yang2016wider}
S.~Yang, P.~Luo, C.-C. Loy, and X.~Tang, ``Wider face: A face detection
  benchmark,'' in \emph{Proceedings of the IEEE conference on computer vision
  and pattern recognition}, 2016, pp. 5525--5533.

\bibitem{everingham2010pascal}
M.~Everingham, L.~Van~Gool, C.~K. Williams, J.~Winn, and A.~Zisserman, ``The
  pascal visual object classes (voc) challenge,'' \emph{International journal
  of computer vision}, vol.~88, no.~2, pp. 303--338, 2010.

\bibitem{bodla2017soft}
N.~Bodla, B.~Singh, R.~Chellappa, and L.~S. Davis, ``Soft-nms—improving
  object detection with one line of code,'' in \emph{Computer Vision (ICCV),
  2017 IEEE International Conference on}.\hskip 1em plus 0.5em minus
  0.4em\relax IEEE, 2017, pp. 5562--5570.

\bibitem{redmon2018yolov3}
J.~Redmon and A.~Farhadi, ``Yolov3: An incremental improvement,'' \emph{arXiv
  preprint arXiv:1804.02767}, 2018.

\end{thebibliography}

\end{document}